%% file: main.tex
\documentclass{article}
     
\usepackage[final]{neurips_2021}
\usepackage[utf8]{inputenc} 
\usepackage[T1]{fontenc}    
\usepackage{hyperref}       
\usepackage{url}            
\usepackage{booktabs}       
\usepackage{amsfonts}       
\usepackage{nicefrac}       
\usepackage{microtype}      
\usepackage{algorithm}
\usepackage{algorithmic}
\usepackage{mathtools}
\usepackage{makecell}
\usepackage{listings}
\usepackage{bm}
\usepackage{booktabs}
\usepackage[table,xcdraw]{xcolor}
\usepackage{natbib}
\usepackage{colortbl}
\usepackage{arydshln}
\usepackage{multirow}
\usepackage{multicol}
\usepackage{xcolor}

\usepackage{enumitem}

\usepackage{enumitem}

\usepackage{graphicx}
\usepackage{xcolor}
\usepackage{float}

\definecolor{codegreen}{rgb}{0,0.6,0}
\definecolor{codegray}{rgb}{0.5,0.5,0.5}
\definecolor{codepurple}{rgb}{0.58,0,0.82}
\definecolor{backcolour}{rgb}{0.95,0.95,0.92}

\lstdefinestyle{mystyle}{
  backgroundcolor=\color{backcolour}, commentstyle=\color{codegreen},
  keywordstyle=\color{magenta},
  numberstyle=\tiny\color{codegray},
  stringstyle=\color{codepurple},
  basicstyle=\ttfamily\footnotesize,
  breakatwhitespace=false,         
  breaklines=true,                 
  captionpos=b,                    
  keepspaces=false,                 
  numbers=left,                    
  numbersep=5pt,                  
  showspaces=false,                
  showstringspaces=false,
  showtabs=false,                  
  tabsize=2
}

\title{ElegantRL-Podracer: Scalable and Elastic Library for Cloud-Native Deep Reinforcement Learning}

\author{
    Xiao-Yang Liu$^1$, Zechu Li$^{1}$\thanks{Equal contribution.}, Zhuoran Yang$^2$, Jiahao Zheng$^3$, Zhaoran Wang$^4$,\\ \textbf{Anwar Walid$^5$\thanks{A. Walid finished this project at Bell labs, before joining Amazon.}, Jian Guo$^6$, Michael I. Jordan$^2$}\\
   $^1$Columbia University; 
   \\$^2$University of California, Berkeley; $^3$Shenzhen Inst. of Advanced Tech.; \\
  $^4$Northwestern University;
  $^5$Amazon \& Columbia University; $^6$IDEA Research.\\
  \texttt{\{xl2427, zl2993\}@columbia.edu; zy6@princeton.edu; jh.zheng@siat.ac.cn} \\
  \texttt{\{zhaoranwang, anwar.i.walid\}@gmail.com; guojian@idea.edu.cn, jordan@cs.berkeley.edu} 
}

\begin{document}

\maketitle

\begin{abstract}

Deep reinforcement learning (DRL) has revolutionized learning and actuation in applications such as game playing and robotic control. The cost of data collection, i.e., generating transitions from agent-environment interactions, remains a major challenge for wider DRL adoption in complex real-world problems. Following a cloud-native paradigm to train DRL agents on a GPU cloud platform is a promising solution. In this paper, we present a scalable and elastic library \textit{ElegantRL-podracer} for cloud-native deep reinforcement learning, which efficiently supports millions of GPU cores to carry out massively parallel training at multiple levels. At a high-level, ElegantRL-podracer employs a tournament-based ensemble scheme to orchestrate the training process on hundreds or even thousands of GPUs, scheduling the interactions between a leaderboard and a training pool with hundreds of pods. At a low-level, each pod simulates agent-environment interactions in parallel by fully utilizing nearly $7,000$ GPU CUDA cores in a single GPU. Our ElegantRL-podracer library features high scalability, elasticity and accessibility by following the development principles of containerization, microservices and MLOps. Using an NVIDIA DGX SuperPOD cloud, we conduct extensive experiments on various tasks in locomotion and stock trading and show that ElegantRL-podracer substantially outperforms RLlib. Our codes are available on GitHub \cite{elegantrl}.

\end{abstract}

\input{NeurIPS_2020_version/Introduction}

\input{NeurIPS_2020_version/RelatedWorks}

\input{NeurIPS_2020_version/Design}

\input{NeurIPS_2020_version/GPU_Podracer}

\input{NeurIPS_2020_version/Evaluation}

\input{NeurIPS_2020_version/Conclusion}

\section*{Acknowledgement}
This research used computational resources of the GPU cloud platform \cite{NVIDIA_SupPod2020} provided by the IDEA Research institute.

\bibliography{main}

\end{document}

%% file: NeurIPS_2020_version/Introduction.tex
\section{Introduction}

Deep reinforcement learning (DRL), which balances the exploration (of uncharted territory) and exploitation (of current information), has revolutionized learning and actuation in applications such as game playing \cite{silver2017mastering} and robotic control \cite{zhang2021reinforcement}. DRL employs a trial-and-error manner to generate transitions from agent-environment interactions, along with the learning procedure. However, the cost of data collection remains a major challenge for wider DRL adoption in real-world problems with complex and dynamic environments. Therefore, a compelling solution is massively parallel training on hundreds or even thousands of GPUs, say millions of GPU cores. 





\begin{figure}[t]
\centering
\includegraphics[width=5in]{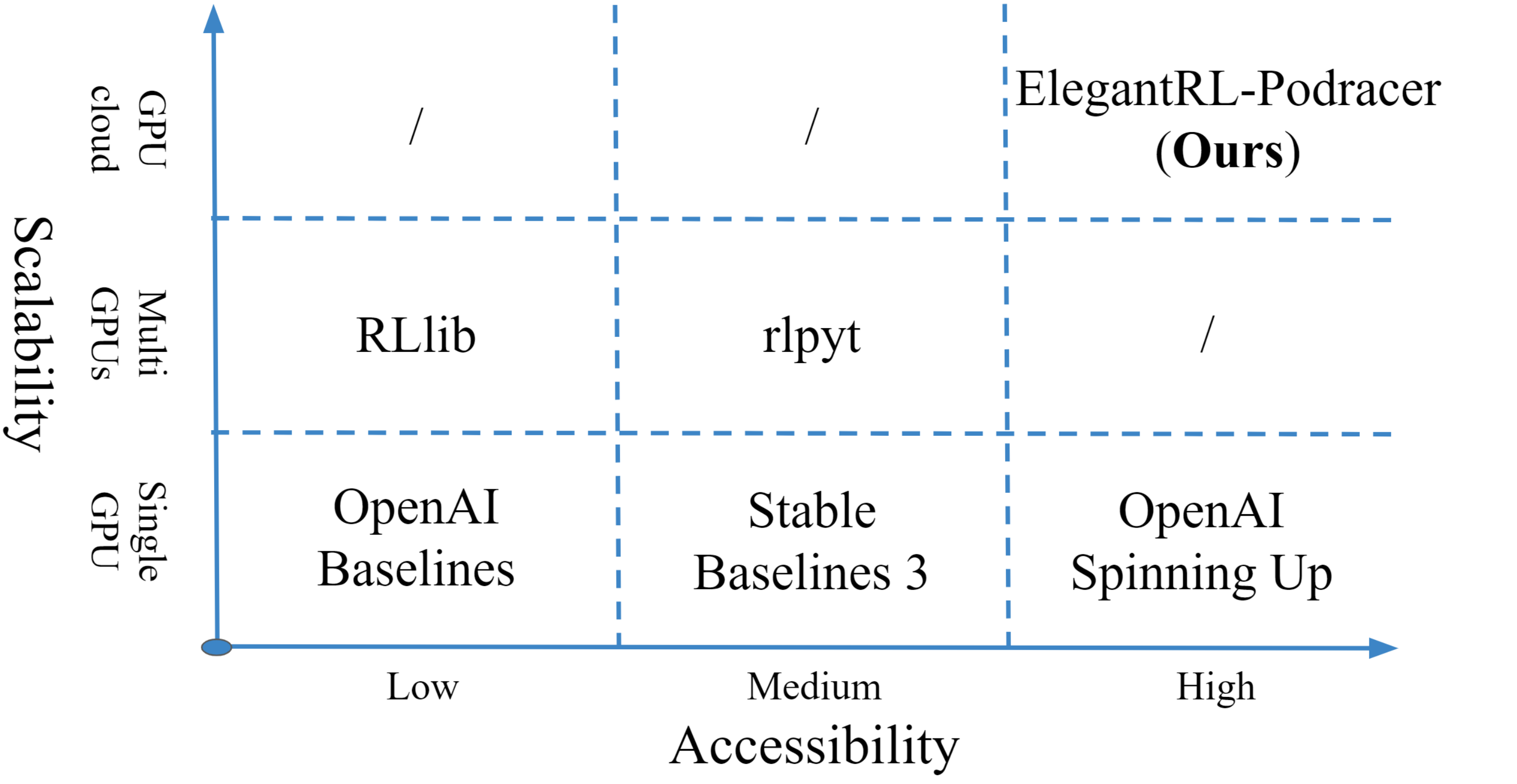}
\vspace{-2mm}
\caption{A comparison of different frameworks/libraries.}
\label{fig_intro}
\vspace{-3mm}
\end{figure}

Existing DRL frameworks are not satisfactory with respect to scalability and accessibility. As shown in Fig. \ref{fig_intro}, OpenAI Baselines \cite{baselines}, Stable Baselines 3 \cite{raffin2019stable} and OpenAI Spinning Up \cite{spinningup} utilize a single GPU, while RLlib \cite{liang2018rllib} and rlpyt \cite{stooke2019rlpyt} can support multiple GPUs. However, there is no existing DRL framework for a cloud with hundreds or even thousands of GPUs. We aim to fully utilize two core techonlogies: 1). GPUDirect technology \cite{NVIDIA_SupPod2020} that provides a path for data to bypass CPUs and travel on “open highways” offered by GPUs, storage, and networking devices; and 2). massively parallel simulations using thousands of GPU cores on a single GPU. On the other hand, the above frameworks, except OpenAI Spinning Up serving an educational purpose, involve a steep learning curve or a lack of customization flexibility, which results in low accessibility.


Scaling out the training process of DRL agents to hundreds or even thousands of GPUs is challenging for researchers and practitioners. The \textit{cloud-native} paradigm aims to scalably and elastically utilize the cloud computing resources. Therefore, we believe it is practically promising to schedule the training of DRL agents by following the cloud-native paradigm, such as employing standardized software stack, e.g., Kubernetes (K8s) \cite{bernstein2014containers}, and adopting core technologies including containers, microservices, continuous integration (CI) and continuous delivery (CD) \cite{Balalaie2016MicroservicesAE, Gannon2017CloudNativeA}.



In this paper, we present a scalable and elastic library \textit{ElegantRL-podracer} for cloud-native deep reinforcement learning, which efficiently utilizes millions of GPU cores to carry out massively parallel training at multiple levels. At a high-level, ElegantRL-podracer employs a tournament-based ensemble scheme to orchestrate the training process on hundreds or even thousands of GPUs, scheduling the interactions between a leaderboard and a training pool with 
hundreds of pods.  At a low-level, each pod simulates agent-environment interactions in parallel by fully utilizing over $7,000$ GPU cores in a single GPU. Our ElegantRL-podracer library features high scalability, elasticity and accessibility by following the development principles of containerization, microservices and MLOps.



Our main contributions are summarized as follows:
\begin{itemize}[leftmargin=*]
    \item We present a scalable and elastic open-source library for cloud-native deep reinforcement learning, \textit{ElegantRL-podracer}, that can utilize millions of GPU cores to train effective DRL agents for complex real-world  problems.
     
    \item To accelerate data collection for efficient exploration, we propose a tournament-based ensemble training scheme and employ massive parallel simulations. 
    
    \item ElegantRL-podracer follows a \textit{cloud-native} paradigm by realizing the development principles of containerization, microservices and MLOps (e.g., continuous integration and continuous delivery), and achieves high accessibility.  
    
    \item Using an NVIDIA DGX SuperPOD cloud \cite{NVIDIA_SupPod2020}, we conduct extensive experiments on various tasks in locomotion and stock trading and show that ElegantRL-podracer substantially outperforms RLlib \cite{liang2018rllib}.

\end{itemize}


The remainder of this paper is organized as follows. Section 2 describes related works. Section 3 presents our design principles. Section 4 describes the ElegantRL-podracer library. In Section 5, we present experimental results. We conclude this paper in Section 6.

%% file: NeurIPS_2020_version/RelatedWorks.tex
\section{Related Works}

We review open-source DRL frameworks/libraries and environment simulation packages.

\subsection{Deep Reinforcement Learning Framework/Libraries}


Many open-source DRL frameworks/libraries have been developed in recent years with varied capabilities. OpenAI Spinning Up \cite{spinningup} and Google Dopamine \cite{castro2018dopamine} are research frameworks for the fast prototyping of DRL algorithms. They both implement numerous DRL algorithms with simple and pedagogical codes. Stable Baseline3 \cite{raffin2019stable} is a stable and efficient DRL library, introducing the parallelism of sampling through vectorized environment. RLlib \cite{liang2018rllib} is a generic DRL library that derives its strength from Ray communication protocols that enable scalable, distributed training. Podracer \cite{hessel2021podracer} from DeepMind is closely related to our ElegantRL-podracer, which also focuses on the efficient usage of large computing resources for training DRL agents. However, it is designed for Google's tensor processing units (TPUs) that are inaccessible to many researchers and practitioners.

\subsection{Simulation Packages}

Environment simulation is a critical component of DRL training, and lots of platforms that provide various task simulations are emerging to close the simulation-to-reality gap. OpenAI Gym \cite{brockman2016openai} is a fundamental simulation toolkit for DRL research, which defines a standard interface for follow-up works. It includes a collection of benchmark problems, e.g., classic control, Atari games, and 2D and 3D robots. MuJoCo \cite{todorov2012mujoco} and Isaac Gym \cite{makoviychuk2021isaac} are two powerful platforms for robotic simulations. MuJoCo \cite{todorov2012mujoco} is a popular physics simulator that efficiently simulates joint contact models. The recently released Isaac Gym \cite{makoviychuk2021isaac} is a high-performance simulation environment for physics. It enables thousands of environments running in parallel on a single GPU. FinRL \cite{liu2020finrl, finrl_podracer_2021} is a new finance-related DRL platform, which simulates various markets as training environments that are built on historical data and live trading APIs.

%% file: NeurIPS_2020_version/Design.tex
\section{Design Principles and Overview}

We aim to develop a user-friendly open-source library that fully exploits cloud resources to train DRL agents. The library emphasizes the following design principles:
\begin{itemize}[leftmargin=*]
    \item \textbf{Scaling-out}: scalability and elasticity.
    \item \textbf{Efficiency}: low communication overhead, massively parallel simulations and robustness of agents.
    \item \textbf{Accessibility}: lightweightness and customization.
\end{itemize}

\begin{figure*}[t]
\centering
\includegraphics[width=5.5in]{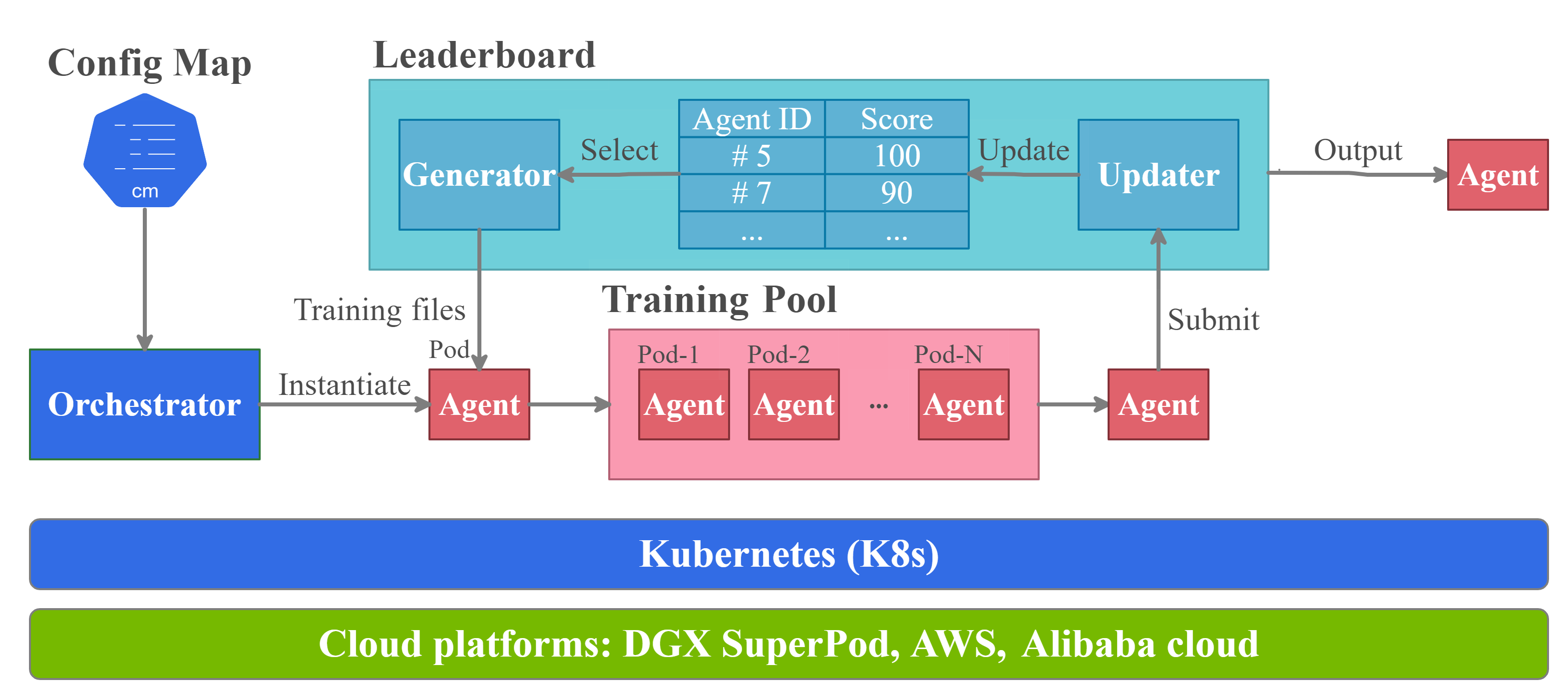}
\vspace{-3mm}
\caption{ElegantRL-podracer employs a tournament-based ensemble training, where a leaderboard is updated by a training pool of pods.}
\label{fig_framework}
\vspace{-3mm}
\end{figure*}

For algorithm design, ElegantRL-podracer employs a \textit{tournament-based ensemble training scheme} to balance exploration and exploitation. In contrast to Evolutionary Strategies (ES) \cite{salimans2017evolution} where a population of agents evolve over generations, our tournament-based ensemble training scheme updates agents asynchronously in parallel, which decouples population evolution and single-agent learning. As shown in Fig. \ref{fig_framework}, the key of the tournament-based ensemble training scheme is the interaction between a \emph{leaderboard} and a \emph{training pool}. The training pool contains hundreds of agents (pods) that 1) are trained in an asynchronous manner, and 2) can be initialized with different DRL algorithms for an ensemble purpose. The leaderboard records the agents with high performance and continually updates as more agents (pods) are trained. 

As shown in Fig. \ref{fig_framework}, the tournament-based ensemble training scheme proceeds as follows:
\begin{enumerate}[leftmargin=*]
    \item An \textit{orchestrator} instantiates a new agent (pod) and put it into a training pool.
    \item A \textit{generator} initializes an agent (pod) with networks and optimizers selected from a leaderboard. The \textit{generator} is a class of subordinate functions associated with the leaderboard, which has different variations to support different evolution strategies.
    \item An \textit{updater} determines whether and where to insert an agent into the leaderboard according to its performance, after a pod has been trained for a certain number of steps or certain amount of time.
\end{enumerate}

For system design, ElegantRL-podracer follows the \textit{cloud-native} paradigm. ElegantRL-podracer achieves containerization by implementing the tournament-based ensemble training scheme as the synergy of microservices. Such a paradigm allows a lightweight usage via simple APIs and a high degree of customization through the flexible cooperation of microservices.

At a high-level, ElegantRL-podracer has the following capabilities to embody our design principles:
\begin{itemize}[leftmargin=*]
    \item \textbf{Asynchronously distributed training} is made possible through a training pool. ElegantRL-podracer can scale out to hundreds or even thousands of computing nodes and elastically adjust the number of agents according to the available computing resources. 
    \item \textbf{Tournament-based ensemble training} is made possible through a leaderboard. Tournament-based training scheme decouples the agent learning and population evolution to achieve low communication overhead between pods. Ensembling many DRL algorithms increases efficiency by exploiting agent robustness and diversity.
    \item \textbf{Cloud-nativity} is made possible with the containerization, microservices, and MLOps adherence. MLOps achieves continuous training/integration/delivery (CT/CI/CD) by exploiting the Kurbernetes (K8s) \cite{bernstein2014containers} software for automated cloud orchestration. 
\end{itemize}



%% file: NeurIPS_2020_version/GPU_Podracer.tex
\section{ElegantRL-Podracer: Scalable and Elastic Cloud-native Library}

In this section, we propose a scalable and elastic cloud-native library, called \textit{ElegantRL-podracer}. We first describe its key components and then present its features.


\subsection{Ensemble Training Using Microservices}

As shown in Fig. \ref{fig_framework}, the ensemble training scheme exploits the synergy of the following microservices: orchestrator, leaderboard (including updater and generator), and agents (pods) in the training pool, where each microservice maps to a  \textit{container}.


\textbf{Orchestrator}: An \textit{orchestrator} monitors the available computing resources and determines the number of pods in the training pool. When K8s signals that the workload is light, the orchestrator generates a set of new pods and insert them into the training pool. When a training objective (i.e., target rewards) is achieved, the orchestrator will terminate the training process. 

\textbf{Leaderboard}: A \textit{leaderboard} records a set of candidate agents with high performance, say cumulative reward. An updater updates the candidate agents to the leaderboard, while a generator instantiates a new pod by referring to candidate agents. The leaderboard may also track other information, such as the covariance matrix, mean, variance, etc, which helps the generator to adaptively allocate computing resources to highly potential candidate agents.
\begin{itemize}[leftmargin=*]
    \item \textbf{Updater}: An \textit{updater} receives a trained agent (pod) and may insert its training files (including actor network, critic network, optimizer parameters, replay buffer (for off-policy algorithms)) into the leaderboard if it has high performance.
    \item \textbf{Generator}: A \textit{generator} generates training files for a newly created pod. The generator may perform a mutation of the candidate agents to increase the diversity.
\end{itemize}

\begin{figure*}[t]
\centering
\includegraphics[width=5.5in]{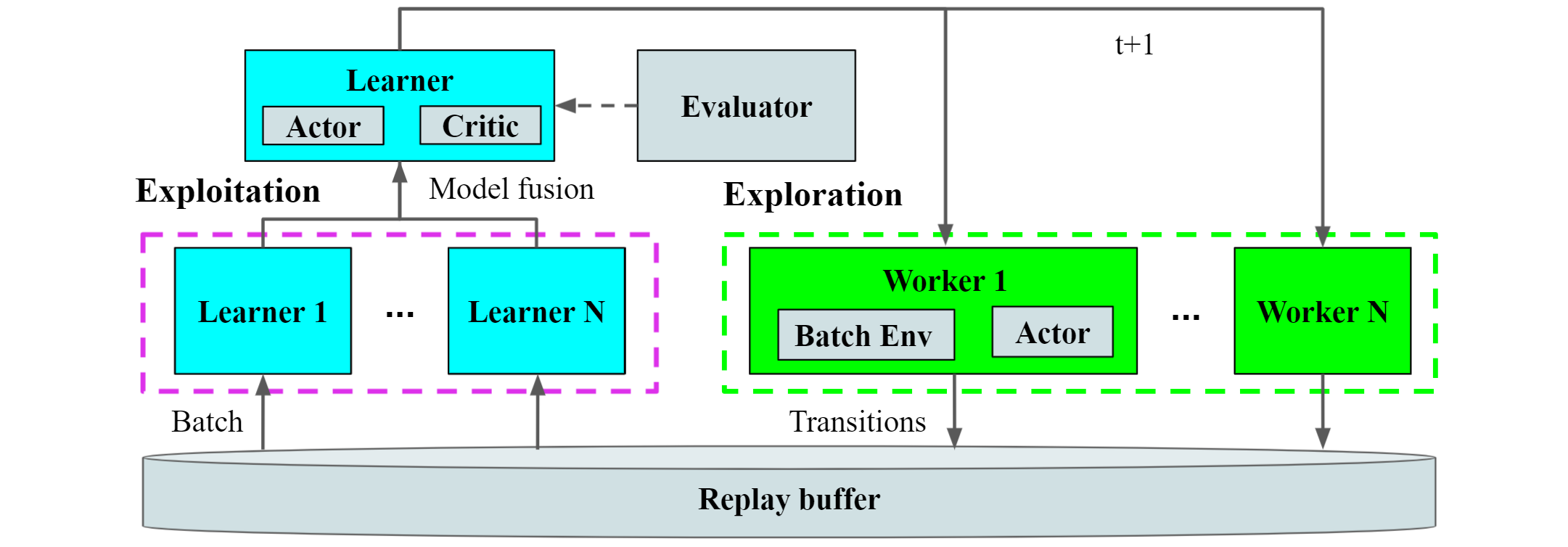}
\vspace{-3mm}
\caption{An agent (pod) is split into three types of microservices: worker, learner and evaluator.}
\label{fig_training}
\vspace{-3mm}
\end{figure*}

\subsection{Agent Learning Using Microservices}

ElegantRL-podracer maps the training process of an agent to a \textit{pod}, which is the smallest deployable unit in K8s \cite{bernstein2014containers}. As shown in Fig. \ref{fig_training}, ElegantRL-podracer separates an agent learning into three microservices: worker, learner and evaluator. 

\textbf{Worker}: A \textit{worker} generates batches of transitions from interactions between the actor and the batched environment. A batched environment consists of multiple independent sub-environments. Each actor collects transitions from the sub-environments of the batched environment in parallel to accelerate the data collection process.

\textbf{Learner}: A \textit{learner} fetches a batch of transitions from the replay buffer to train the neural networks. ElegantRL-podracer trains multiple learners in parallel and fuses the networks by aggregating \textbf{network parameters}. In this way, ElegantRL-podracer experiments much less communication overhead than distributed SGD in RLlib \cite{liang2018rllib}. 

\textbf{Evaluator}: An \textit{evaluator} continuously evaluates a pod and records its performance and corresponding networks during the training process. Commonly used performance metrics are the mean and variance of the episodic reward. Note that the evaluator effectively mitigates the performance loss caused by either overfitting or early stopping.

\subsection{Features of ElegantRL-podracer}

ElegantRL-podracer achieves several features that facilitate the implementation of a lightweight and powerful training process on a GPU cloud.

\textbf{Scalable parallelism}: The multi-level parallelism of ElegantRL-podracer leads to high scalability.
\begin{itemize}[leftmargin=*]
    \item \textbf{Agent parallelism}: The agents in the training pool are parallel, thus can easily scale out to a large number. The asynchronous training of parallel agents can also reduce the frequency of agent-to-agent communication.
    \item \textbf{Learner parallelism}: An agent employs multiple learners to train the neural networks in parallel, and then fuse the networks parameters to obtain a result agent, instead of using distributed SGD. Such a model fusion through network parameters involves a much lower frequency communication as the fusion process only happens at the end of an epoch.
    \item \textbf{Worker parallelism}: An agent utilizes multiple rollout workers to sample transitions in parallel.
\end{itemize}

\textbf{Elastic resource allocation}: The elasticity is critical for cloud-level applications as it helps users adapt to the changes in cloud resources and prevent over-provisioning and under-provisioning of resources \cite{Mell2011TheND, Herbst2013ElasticityIC}. ElegantRL-podracer can elastically allocate the number of agents (pods) by employing an orchestrator to monitor the available computing resources and the current training status.

\textbf{Cloud-oriented optimizations}: ElegantRL-podracer co-locates microservices on GPUs to accelerate the parallel computation on both data collection and model training. For the data transfer and storage, ElegantRL-podracer represents data as tensors to speedup the communication and allocates the shared replay buffer on the contiguous memory of GPUs to increase the addressing speed.

\textbf{Continuous training (CT) pipeline}: Continuous training, which is a part of the MLOps practice, seeks to automatically and continuously retrain the model to adapt to changes that might occur in the data. ElegantRL-podracer performs the CT of a DRL agent by automating a lightweight DRL training pipeline, which is composed of microservices. Users can conduct different experiments and hyper-parameter search by modifying workers, learners and other microservices.

\textbf{Continuous integration/delivery (CI/CD)}: ElegantRL-podracer enables a robust CI/CD for users to explore new ideas by modifying existing microservices or build new microservices. The microservices are loosely coupled, such that the change of one microservice will not break existing ones. Also modularity allows for more comprehensive search over the experiment space, e.g., instead of designing one experiment at a time, we could theoretically be able to test $c_1 \times c_2 \times \cdot \times c_n$ experiments in an automated fashion, where $n$ is the number of components for a DRL algorithm and $c_i$ is the number of optional microservices for component $i$, $i=1,...,n$:
\begin{itemize}[leftmargin=*]
    \item \textbf{Environment variation}: ElegantRL-podracer supports any environment written in gym-style and provides a class \textit{PreprocessVecEnv} to convert a normal environment to a batch mode. 
    \item \textbf{Algorithm variation}: ElegantRL-podracer can realize different DRL algorithms through combinations of worker and learner variants. Currently, ElegantRL-podracer supports fine-tuned standard DRL algorithms, including DQN-series \cite{Mnih2013PlayingAW, Hasselt2016DeepRL}, DDPG \cite{Lillicrap2016ContinuousCW}, TD3 \cite{Fujimoto2018AddressingFA}, SAC \cite{Haarnoja2018SoftAO}, and PPO \cite{Schulman2017ProximalPO}. New algorithms may be used as long as they adhere to the agent interface.
    \item \textbf{Evolution variation}: ElegantRL-podracer allows users to customize the evaluator, updater and generator in the leaderboard to decide \textit{how to evaluate}, \textit{where to update}, and \textit{what to generate}.
\end{itemize}

%% file: NeurIPS_2020_version/Evaluation.tex
\section{Performance Evaluation}
\label{evaluation}

In the section, we describe the cloud platform in our experiments and the performance of ElegantRL-podracer on various tasks from robotic control to stock trading task.

\subsection{Experiment Platform: GPU Cloud}

All experiments were executed using NVIDIA DGX-2 servers \cite{choquette2021nvidia} in a DGX SuperPOD cloud \cite{NVIDIA_SupPod2020}, a cloud-native infrastructure. Each DGX-2 server contains $8$ A100 GPUs and $320$ GB GPU memory in total, and also has $128$ CPU cores of Dual AMD Rome 7742 running at 2.25GHz. Among the $8$ A100 GPUs in one DGX-2 server, any two A100 GPUs are connected with each other through $12$ NVLinks, providing $600$ Gbps of full-duplex bandwidth \cite{choquette2021nvidia}.




\subsection{Robotic Control Tasks}
\label{robot}

\begin{figure*}[th]
    \centering
    \includegraphics[width=1.2in, height=1.2in]{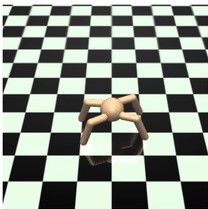}
    \hspace{0.3cm}
    \includegraphics[width=1.2in, height=1.2in]{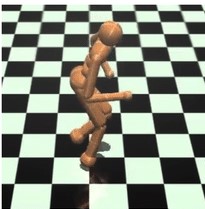}
    \hspace{0.3cm}
    \includegraphics[width=1.2in, height=1.2in]{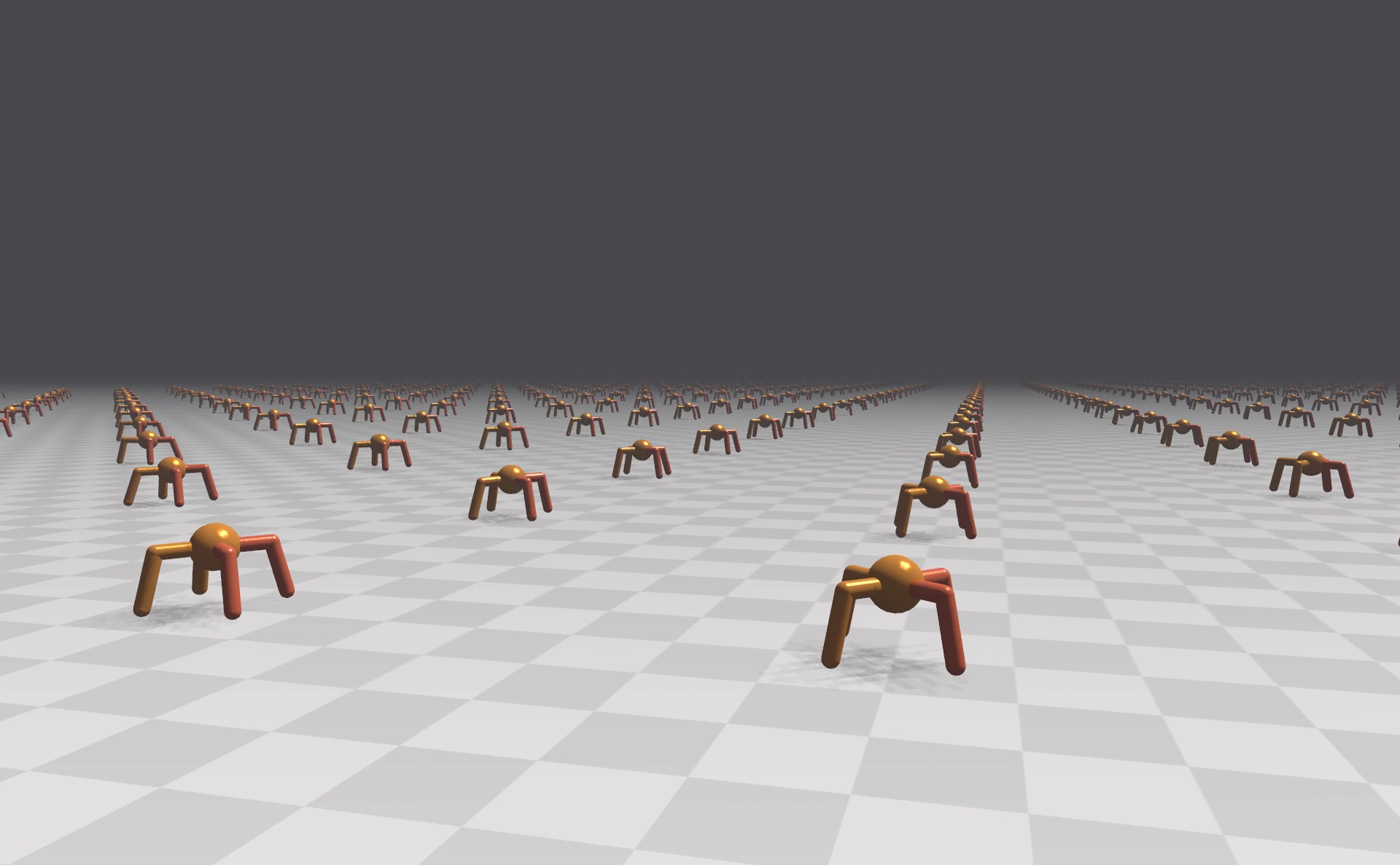}
    \hspace{0.3cm}
    \includegraphics[width=1.2in, height=1.2in]{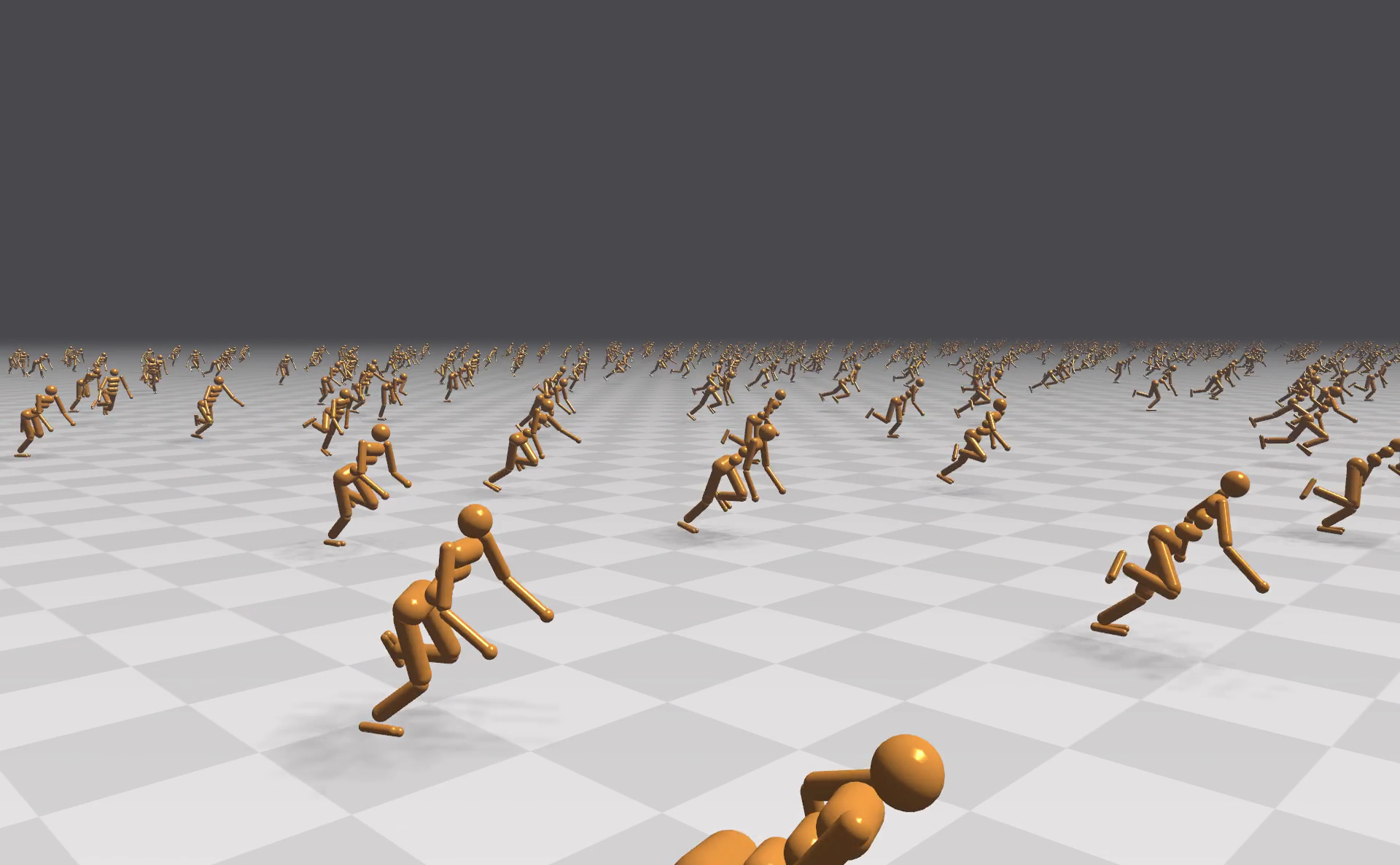}
    \vspace*{-1mm}
    \caption{Snapshots of robotic control environments. From left to right, the ant and humanoid tasks from MuJoCo \cite{todorov2012mujoco}, and the ant and humanoid tasks from Isaac Gym \cite{makoviychuk2021isaac}.}
    \vspace*{-3mm}
    \label{fig_robotic}
\end{figure*}

\textbf{Ant and humanoid} are two canonical robotic control tasks that simulate an ant and a humanoid, respectively, where each task has both MuJoCo \cite{todorov2012mujoco} version and Isaac Gym \cite{makoviychuk2021isaac} version, as shown in Fig. \ref{fig_robotic}. The ant task is a simple environment to simulate due to its stability in the initial state, while the humanoid task is often used as a testbed for locomotion learning. Even though the implementations of MuJoCo \cite{todorov2012mujoco} and Isaac Gym \cite{makoviychuk2021isaac} are slightly different, the objective of both is to have the agent move forward as fast as possible. The state space, action space and reward function are given in Table \ref{envs}. We select the same tasks from the two platforms in order to show that 1) ElegantRL-podracer can support different simulator platforms, and 2) the potential of massively parallel simulations in the DRL training by comparing the CPU-based MuJoCo \cite{todorov2012mujoco} with the GPU-based Isaac Gym \cite{makoviychuk2021isaac}.

\vspace*{-3mm}

\begin{table}[]
\small
\renewcommand{\arraystretch}{1.2}
\centering
\begin{tabular}{|l|l|l|l|m{200pt}<{\centering}|}
   \hline   
    \textbf{Tasks} & \textbf{State space $\mathcal{S}$} & \textbf{Action space $\mathcal{A}$} & \textbf{Reward $r(s, a, s')$} \\
   \hline
    \multirow{4}{2.4cm}{Ant \cite{todorov2012mujoco, makoviychuk2021isaac}} & Body height and rotation & 8 controllable joints & Alive bonus \\
    \cdashline{2-4}[0.8pt/2pt]
    & Velocity and angular velocity & & Running speed \\
    \cdashline{2-4}[0.8pt/2pt]
    & Joint angles & & Standing and Heading\\
    \cdashline{2-4}[0.8pt/2pt]
    & Forces, etc. & & Contact forces\\
    \hline
     \multirow{4}{2.4cm}{Humanoid \cite{todorov2012mujoco, makoviychuk2021isaac}} & Body height and rotation & 17 joints for MuJoCo & Alive bonus \\
    \cdashline{2-4}[0.8pt/2pt]
    & Velocity and angular velocity & 21 joints for Isaac gym & Running speed\\
    \cdashline{2-4}[0.8pt/2pt]
    & Joint angles & & Standing and Heading\\
    \cdashline{2-4}[0.8pt/2pt]
    & Forces, etc. & & Contact forces\\
    \hline
     \multirow{3}{2.4cm}{Stock trading \cite{liu2020finrl}} & Balance, Shares & Buy & Change of account value\\
    \cdashline{2-4}[0.8pt/2pt]
    & Close prices & Sell &\\
    \cdashline{2-4}[0.8pt/2pt]
    & Technical indicators & Hold & \\
   \hline
   
\end{tabular}
\vspace*{2mm}
\caption{The state space, action space and reward function of ant, humanoid and stock trading tasks.}
\vspace*{-5mm}
\label{envs}
\end{table}


\textbf{Compared methods}: On one DGX-2 server, we compare ElegantRL-podracer with RLlib \cite{liang2018rllib}, since both support multiple GPUs. ElegantRL-podracer used PPO \cite{Schulman2017ProximalPO} from ElegantRL \cite{elegantrl}, while in RLlib \cite{liang2018rllib} we used the Decentralized Distributed Proximal Policy Optimization (DD-PPO) \cite{Wijmans2020DDPPOLN} algorithm that scales well to multiple GPUs. For fair comparison, we keep all adjustable parameters and computing resources the same, such as the depth and width of neural networks, total training steps/time, number of workers, and GPU and CPU resources. Specifically, we use a batch size of 1024, learning rate of 0.001, and a replay buffer size of 4096 across tasks.

\textbf{Performance metrics}: We employ two different metrics to evaluate the agent's performance:
\begin{itemize}[leftmargin=*]
    \item \textbf{Episodic reward vs. training time (wall-clock time)}: we measure the episodic reward at different training time, which can be affected by the convergence speed, communication overhead, scheduling efficiency, etc.
    \item \textbf{Episodic reward vs. training step}: from the same testings, we also measure the episodic reward at different training steps. This result can be used to investigate the massive parallel simulation capability of GPUs, and also check the algorithm's performance.
\end{itemize}

During the training process, we evaluate the agent $10$ times to obtain $10$ episodic rewards and report the average episodic reward and standard deviation.

\begin{figure*}[]
    \centering
    \includegraphics[width=0.245\textwidth]{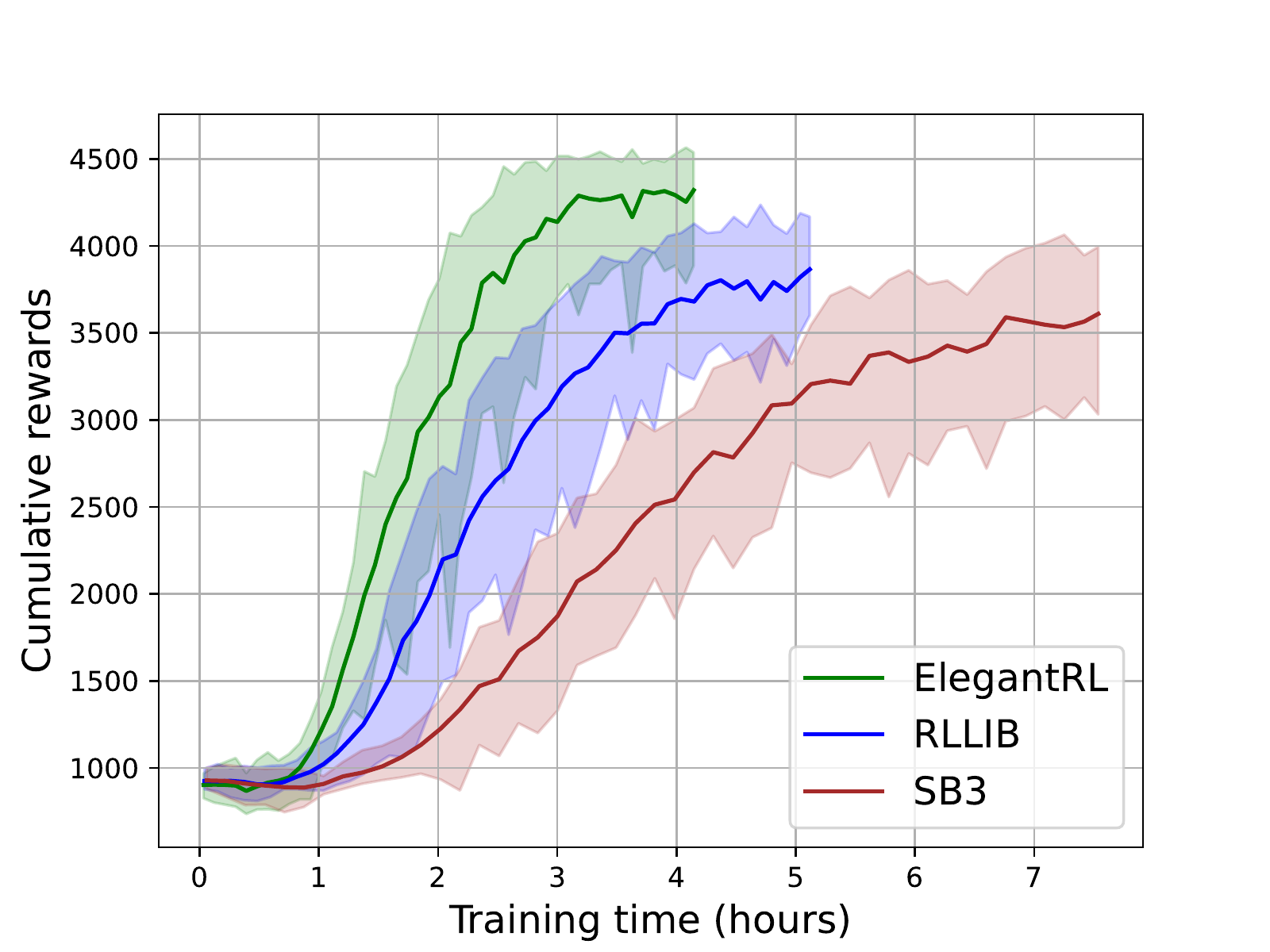}
    \hspace{-0.16cm}
    \includegraphics[width=0.245\textwidth]{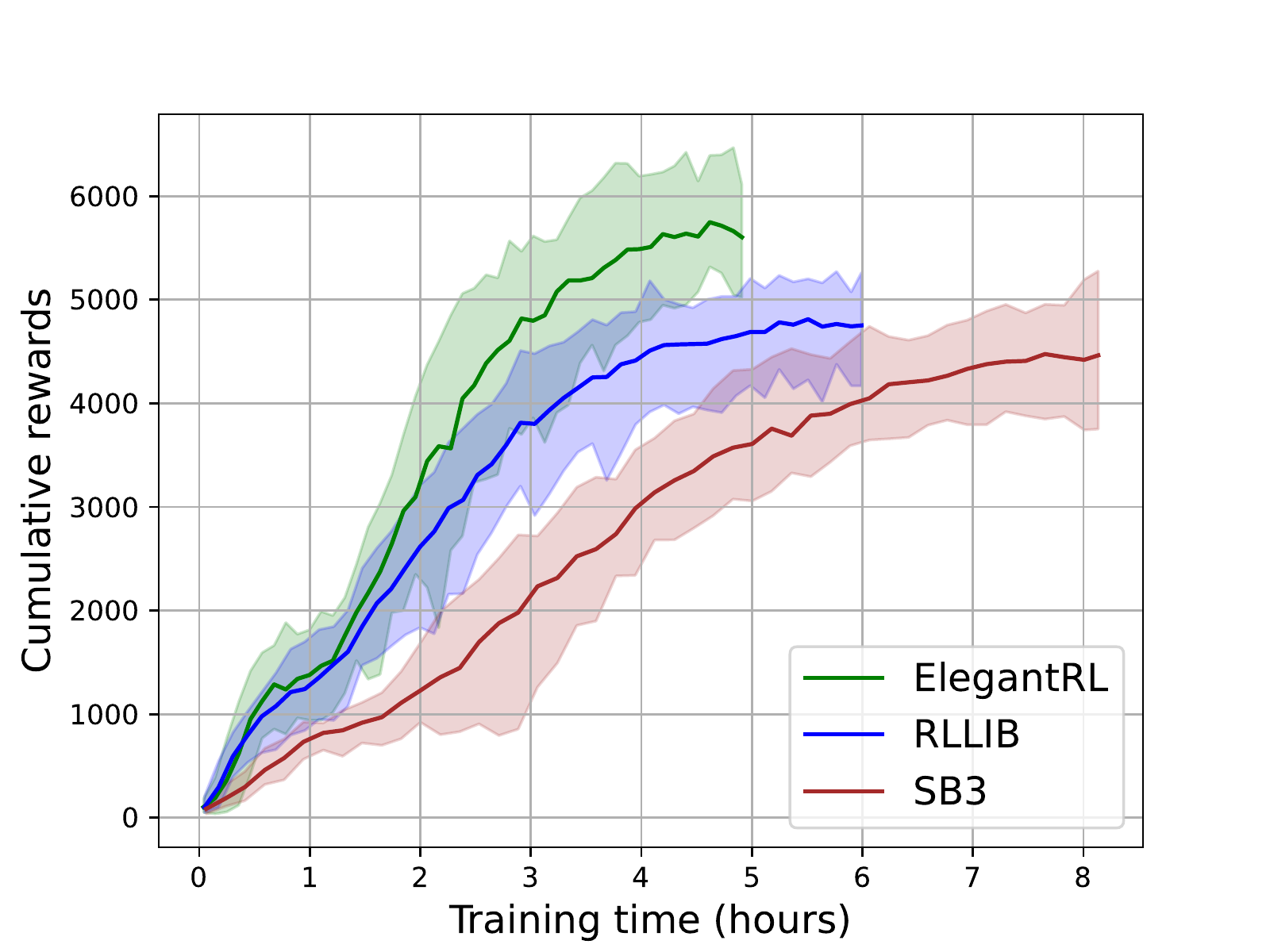}
    \hspace{-0.16cm}
    \includegraphics[width=0.255\textwidth]{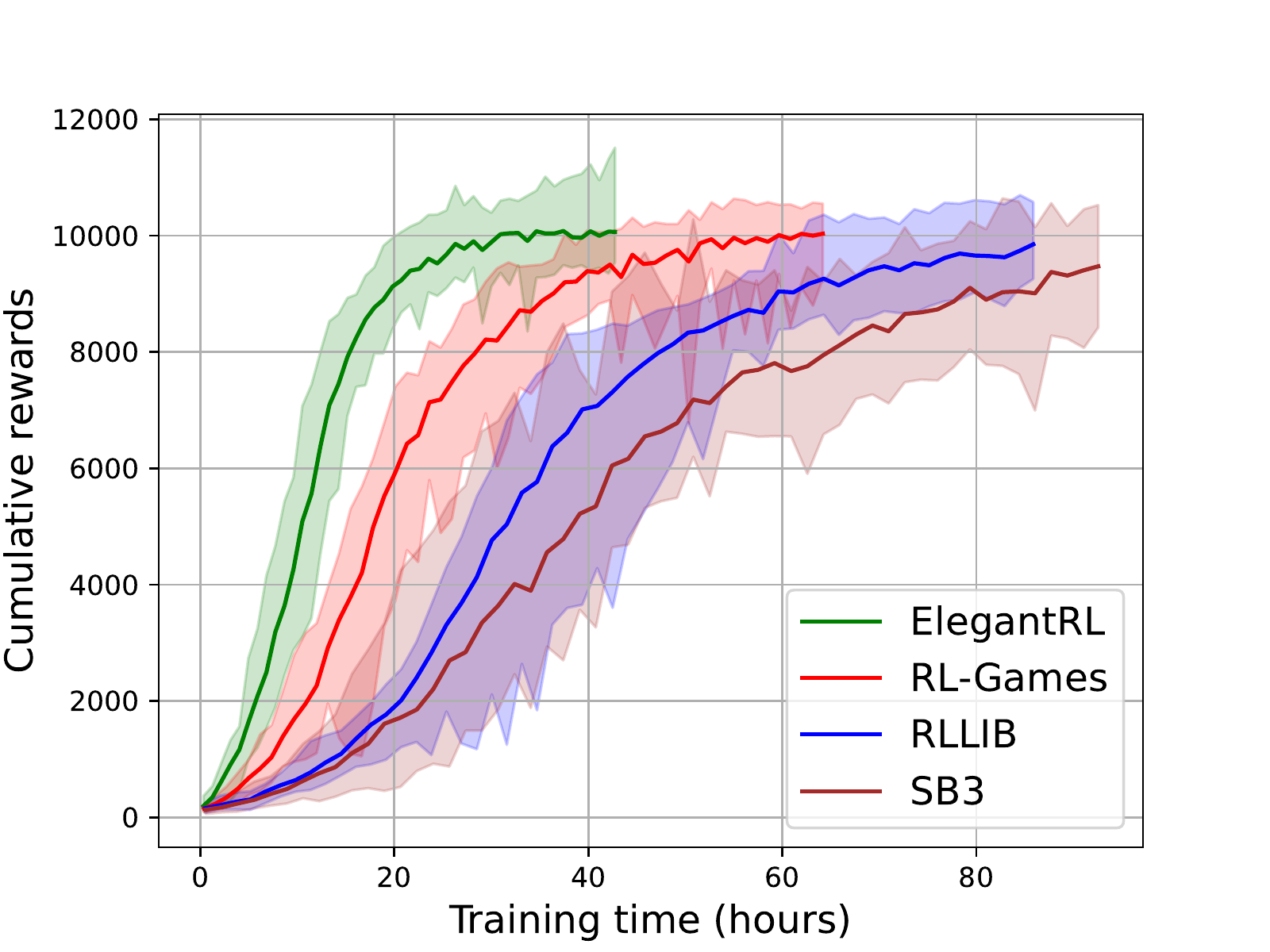}
    \hspace{-0.2cm}
    \includegraphics[width=0.255\textwidth]{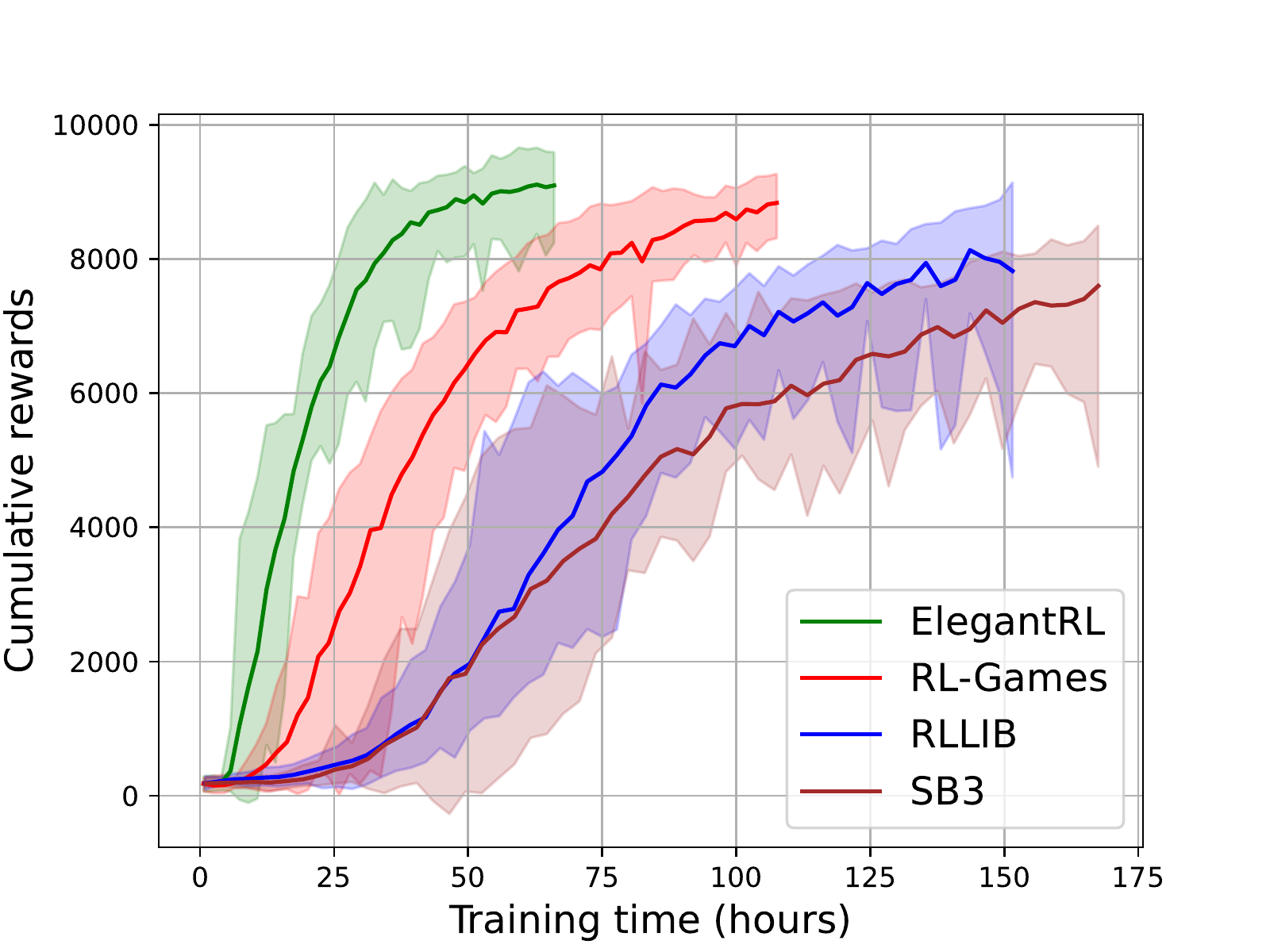}
   \vspace{-5mm}
   \caption{Episodic reward vs. training time (wall-clock time) for the four tasks in Fig. \ref{fig_robotic}. }
   \label{fig_robotic_time}
\end{figure*}

For the four tasks in Fig. \ref{fig_robotic}, we terminate the training processes at $8,000$s ($\approx 2.2$ hours), $32,000$s ($\approx 8.9$ hours), $25,000$s ($\approx 6.9$ hours) and $9,000$s ($\approx 2.5$ hours), respectively. As shown in Fig. \ref{fig_robotic_time}, we can see that given the same training time, ElegantRL-podracer achieves substantially higher episodic rewards than RLlib. For Isaac Gym in particular, the corresponding episodic rewards have been nearly doubled. In the task ant (Isaac Gym), RLlib needs approximately $7.0$ hours to achieve a reward $9,000$, while Elegant-podracer only needs approximately $1.4$ hours to get the same reward, which is $5\times$ faster. 

\begin{figure*}[]
    \centering
    \includegraphics[width=0.245\textwidth]{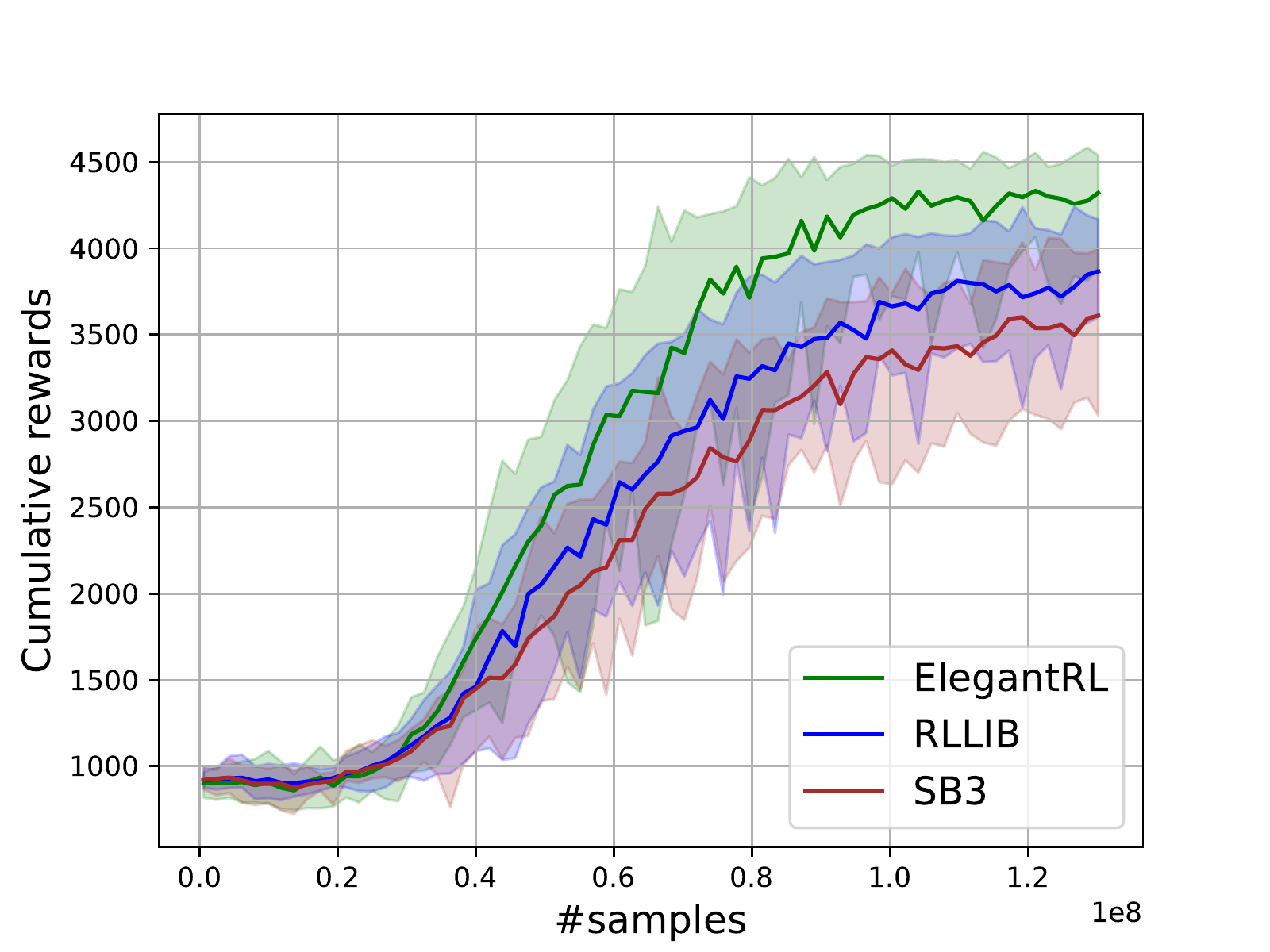}
    \hspace{-0.16cm}
    \includegraphics[width=0.245\textwidth]{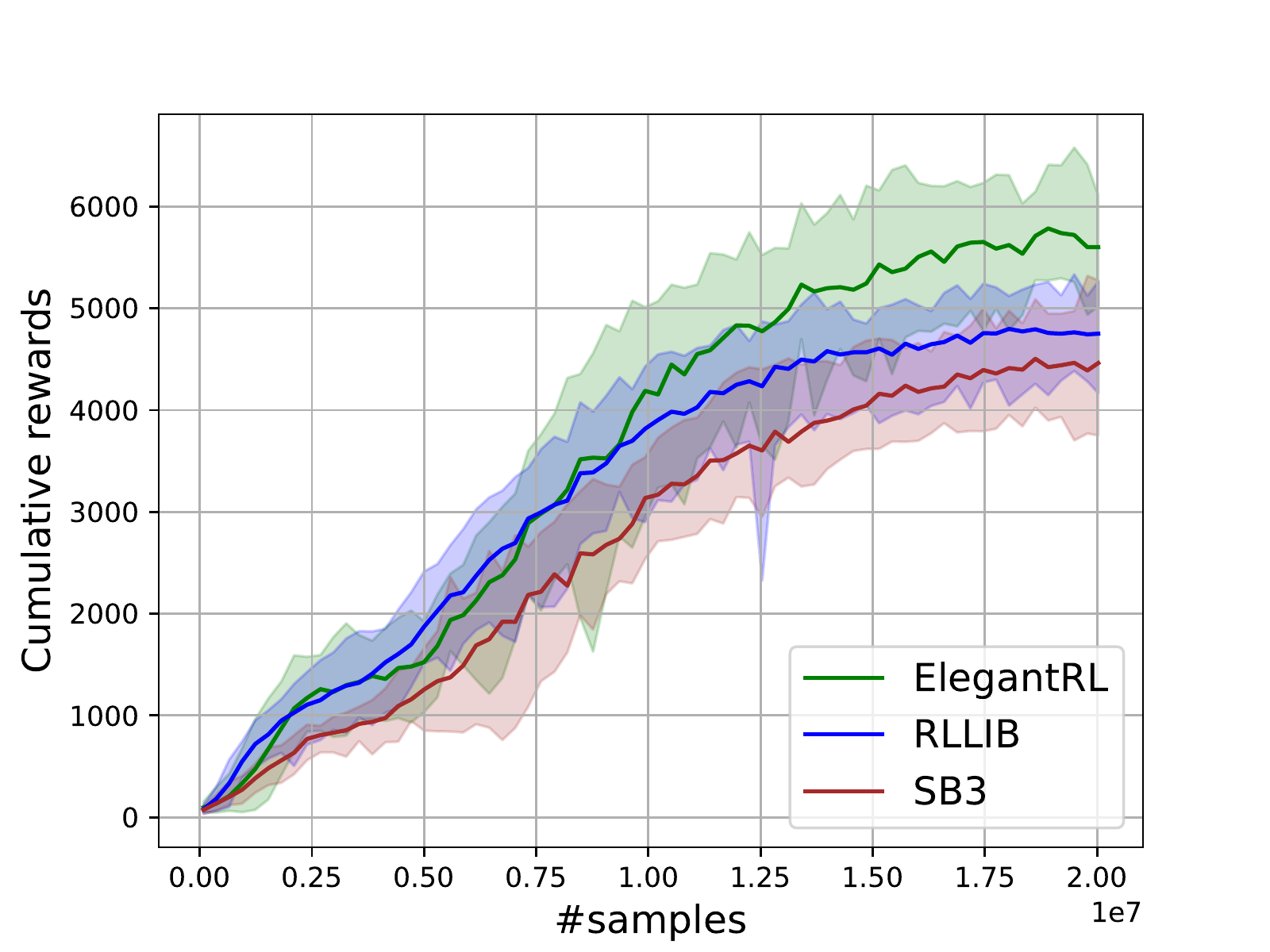}
    \hspace{-0.16cm}
    \includegraphics[width=0.255\textwidth]{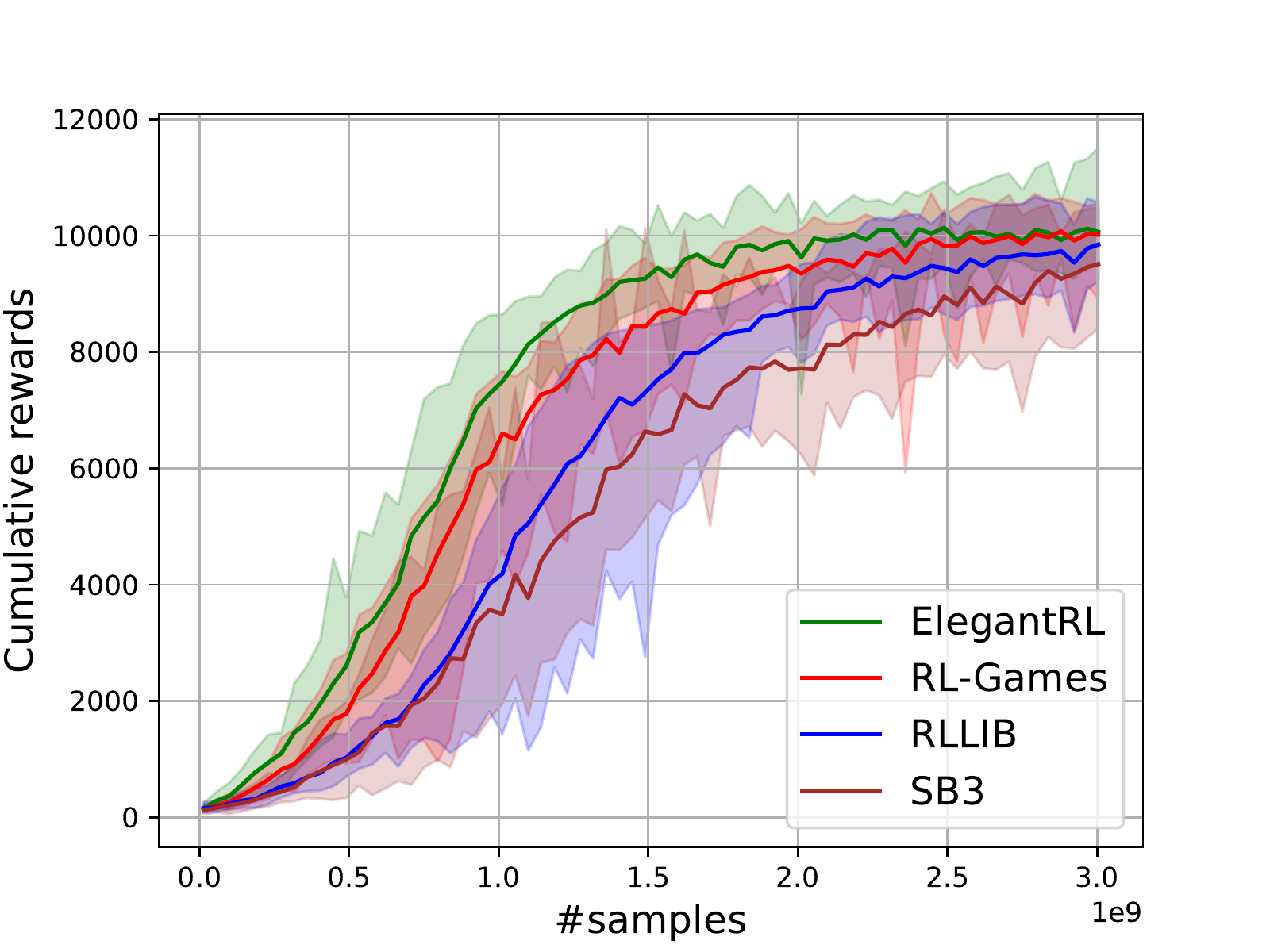}
    \hspace{-0.2cm}
    \includegraphics[width=0.255\textwidth]{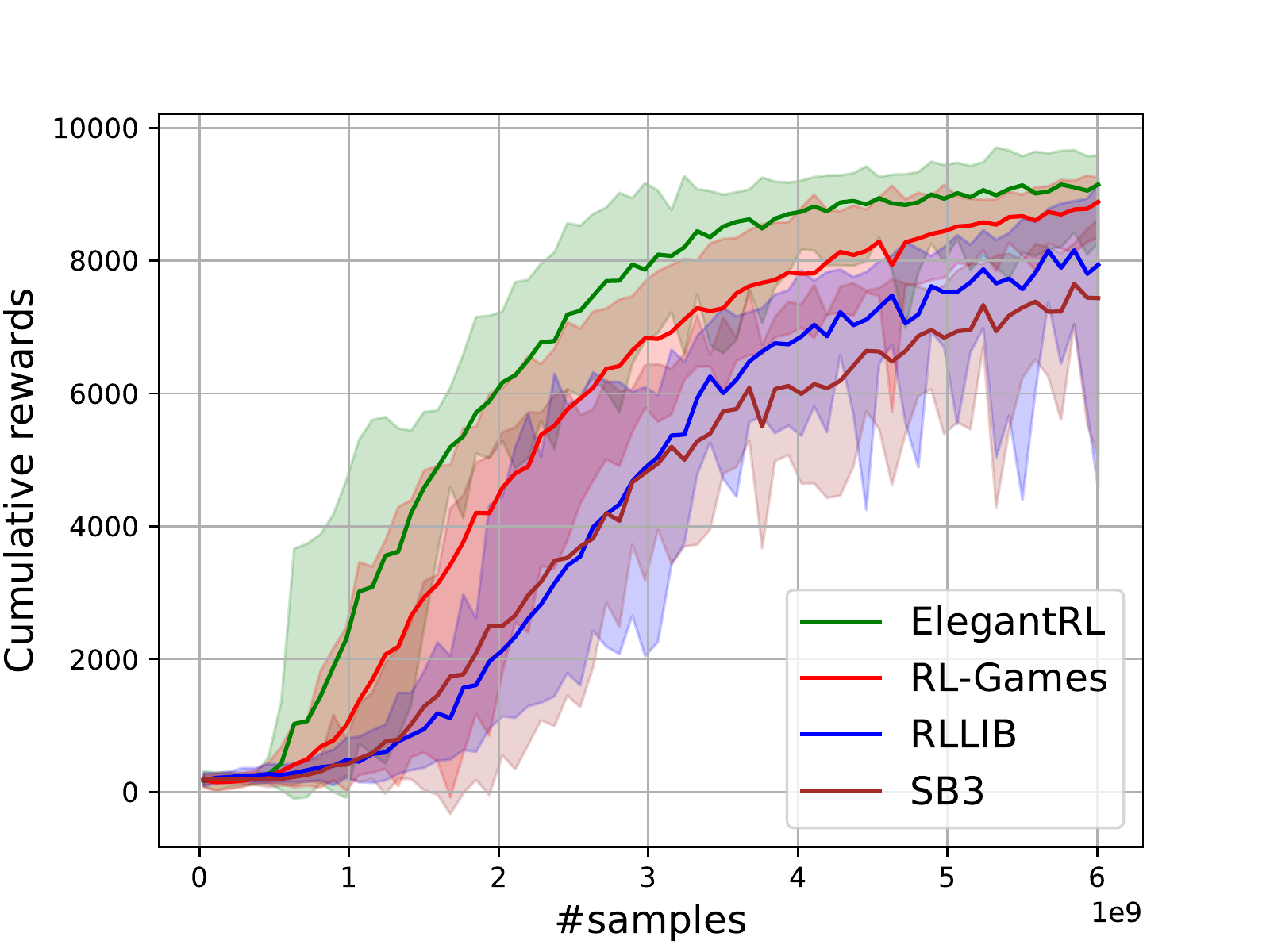}
    \vspace{-5mm}
  \caption{Episodic reward vs. training steps for the four tasks in Fig. \ref{fig_robotic}.}
  \label{fig_robotic_step}
  \vspace{-5mm}
\end{figure*}

We run $2.0 \times 10^7$ steps, $2.5 \times 10^7$ steps,  $4 \times 10^8$ steps and $8 \times 10^8$ steps, respectively. Take a closer look at Fig. \ref{fig_robotic_step}, we can see that ElegantRL-podracer achieves higher episodic rewards than RLlib in all four tasks. A possible reason is the tournament-based ensemble training scheme guide a population of agents update toward a direction with higher rewards.


\subsection{Stock Trading Task}

Finance is a promising and challenging real-world application of DRL algorithms. We apply ElegantRL-podracer to a stock trading task as an example  \cite{liu2020finrl,finrl_podracer_2021} to show its potential in quantitative finance. 

\textbf{Stock trading task}: we aim to train a DRL agent that decides \textit{where to trade}, \textit{at what price and what quantity} in a stock market, thus the objective of the problem is to maximize the expected return and minimize the risk. We model the stock trading task as a Markov Decision Process (MDP) as in FinRL \cite{liu2020finrl,finrl_podracer_2021}, where the state space, action space and reward function are given in Table \ref{envs}.

\textbf{Data pre-processing}: We select the NASDAQ-100 constituent
stocks as our stock pool and use the minute-level dataset for our experiment. For the data preparation, we download the raw data from the Compustat database through the Wharton Research Data Services (WRDS) \cite{wrds}. Next, we process it to an open-high-low-close-volume (OHLCV) format and extract technical indicators. Finally, we follow a training-backtesting pipeline and split the dataset into two sets: the data from 01/01/2016 to 05/25/2020 for training, and the data from 05/26/2020 to 05/26/2021 for backtesting.

\textbf{Performance metrics}: We evaluate trading performance and training performance, respectively. Five common metrics are used to quantify the trading performance:
\begin{itemize}[leftmargin=*]
    \item \textbf{Cumulative return}: subtracting the initial value from the final portfolio value, then dividing by the initial value.  
    \item \textbf{Annual return and volatility}: geometric average return in a yearly sense, and the deviation.
    \item \textbf{Maximum drawdown}: the maximum observed loss from a historical peak to a trough of a portfolio, before a new peak is achieved. It is an indicator of downside risk over a time period.
    \item \textbf{Sharpe ratio}: the average return earned in excess of the risk-free rate per unit of volatility.
    \item \textbf{Calmar ratio}: the fund's average compounded annual rate of return versus its maximum drawdown.
\end{itemize}

For the training performance, we use the metric, episodic reward vs training time (wall-clock time), in Section \ref{robot}. We record the required training time for reaching a specific cumulative return. 

We reserve a time period not used for training but only testing to evaluate generalization performance for the stock trading problem. Since the agent cannot access the testing dataset during the training, we store the model snapshots at different training times, say every 100 seconds. Later, we use each snapshot model to perform inference on the testing dataset to obtain the cumulative return. 

\textbf{Compared methods}: We compare ElegantRL-podracer with RLlib \cite{liang2018rllib} with the same setup in Section \ref{robot}. Invesco QQQ ETF is the benchmark to represent the market performance. There are in total $80$ A100 GPUs assigned to our usage.

\begin{figure*}
    \begin{minipage}{\textwidth}
    \centering
    \includegraphics[width=2.7in]{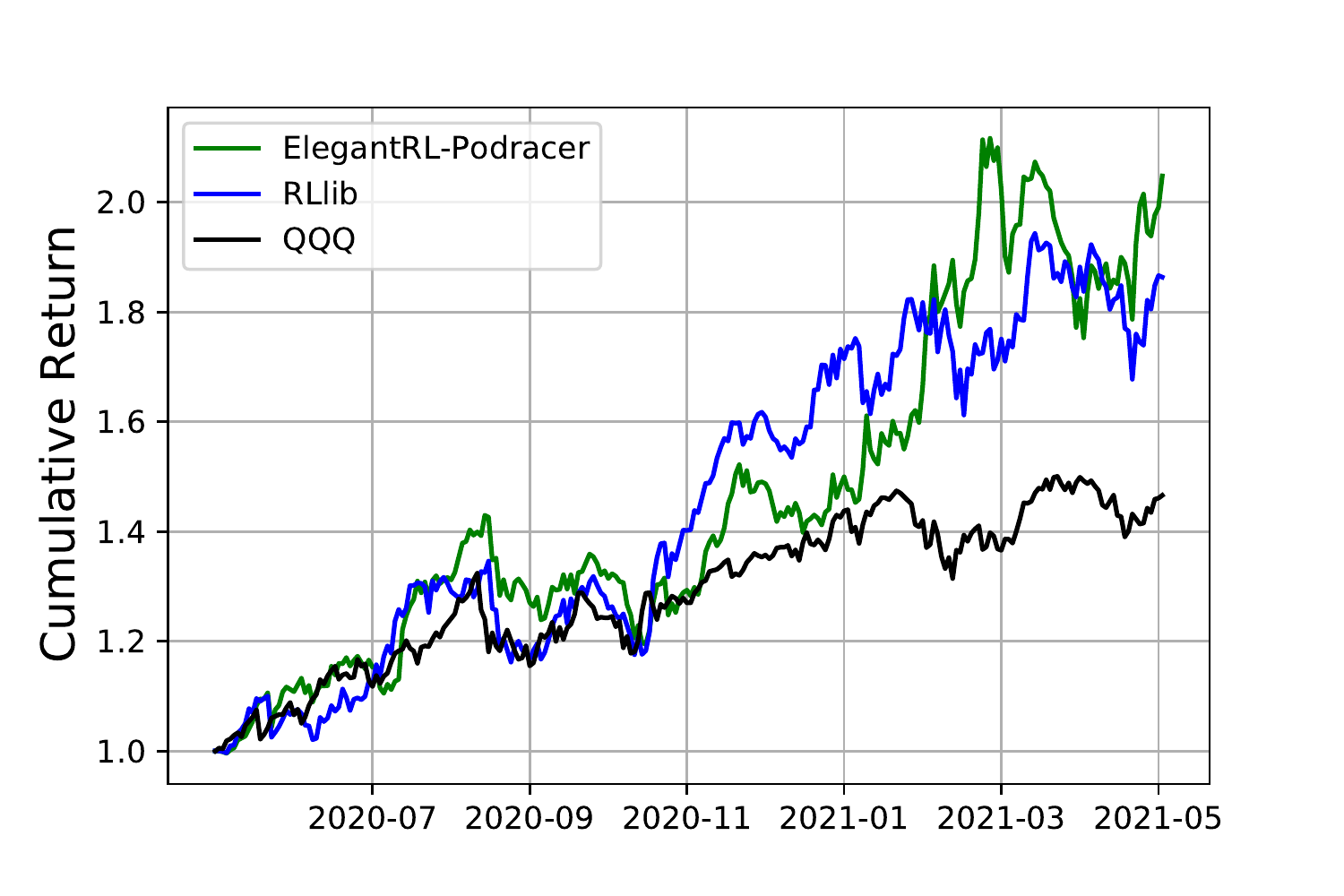}
    \hspace{-1mm}
    \includegraphics[width=2.7in]{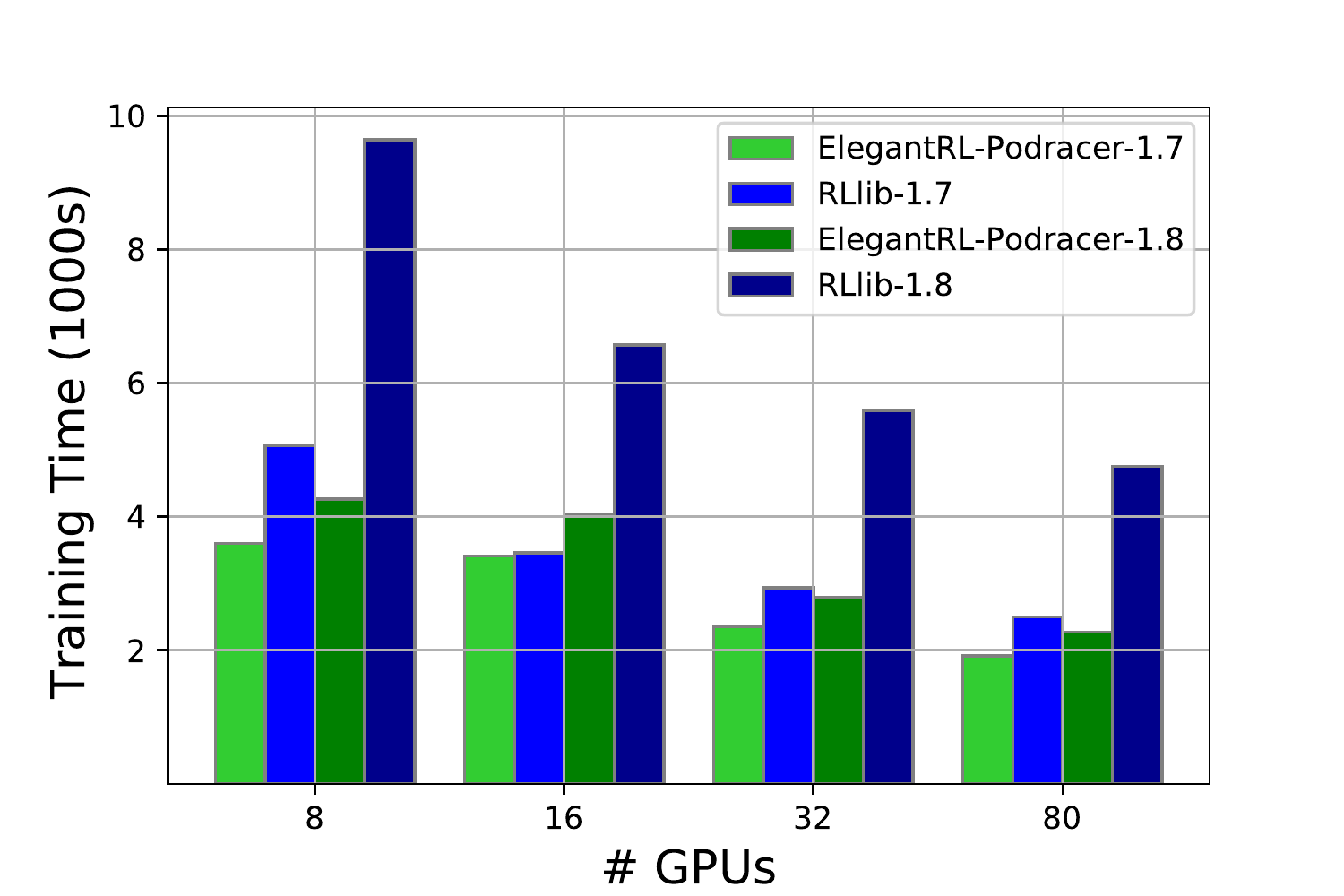}
    \end{minipage}
    \vspace*{-2mm}
   \caption{Left: cumulative return on minute-level NASDAQ-100 constituents stocks (initial capital $\$1,000,000$, transaction cost $0.2\%$). Right: training time (wall-clock time) for reaching cumulative rewards 1.7 and 1.8, using the model snapshots of ElegantRL-podracer and RLlib \cite{liang2018rllib}.}
   \label{fig_nas}
   \vspace*{-3mm}
\end{figure*}

From Fig. \ref{fig_nas}, all DRL agents can achieve a better performance than the market benchmark with respect to the cumulative return, demonstrating the algorithm's effectiveness. According to Table \ref{tab:nas}, we observe that ElegantRL-podracer has cumulative return 104.743\%, annual return 103.591\%, and Sharpe ratio 2.20, which outperforms RLlib substantially. However, ElegantRL-podracer is not as stable as RLlib during the backtesting period: it achieves an annual volatility 35.357\%, max. drawdown -17.187\%, and Calmar ratio 6.02. There are two possible reasons to account for such an instability: 1) the reward design in the stock trading environment is mainly related to the cumulative return, thus leading the agent to take less care of the risk; 2) ElegantRL-podracer holds a large amount of funds around 2021-03, as shown in Fig. \ref{fig_nas}, which naturally leads to a larger slip.

We compare the training performance on a varying number of GPUs, i.e., 8, 16, 32, and 80. We measure the required training time to obtain two cumulative returns 1.7 and 1.8, respectively. In Fig. \ref{fig_nas}, both ElegantRL-podracer and RLlib \cite{liang2018rllib} requires less training time to achieve the same cumulative return as the number of GPUs increases, which directly demonstrates the advantage of cloud computing resources on the DRL training. For ElegantRL-podracer with 80 GPUs, it requires (1900s, 2200s) to reach cumulative returns of 1.7 and 1.8. ElegantRL-podracer with 32 and 16 GPUs need (2400s, 2800s) and (3400s, 4000s) to achieve the same cumulative returns. It demonstrates the high scalability of ElegantRL-podracer and the effectiveness of our cloud-oriented optimizations. For the experiments using RLlib \cite{liang2018rllib}, increasing the number of GPUs does not lead to much speed-up. 

\begin{table}[t]
		\centering
		\begin{tabular}{|l|c|c|c|c|c|c|}\hline
			 & \makecell{Cumul. \\ return} & \makecell{Annual \\ return} & \makecell{Annual \\ volatility} & \makecell{Max. \\ drawdown} & \makecell{Sharpe \\ ratio} & \makecell{Calmar \\ ratio} \\
			\hline
			ElegantRL-podracer & \textbf{104.743\%} & \textbf{103.591\%} & 35.357\% & -17.187\% & \textbf{2.20} & 6.02\\
			
			RLlib \cite{liang2018rllib} & 86.274\% & 85.364\% & 34.319\%  & -13.689\% & 1.98 & \textbf{6.24}\\

			Invesco QQQ ETF & 46.586\% & 46.146\% & \textbf{23.39\%} & \textbf{-12.749\%} & 1.75 & 3.62\\
			\hline
		\end{tabular}
		\vspace*{1mm}
		\caption{Stock trading performance on NASDAQ-100 constituent stocks with minute-level data.}
		\label{tab:nas}
		\vspace{-3mm}
\end{table}



%% file: NeurIPS_2020_version/Conclusion.tex
\section{Discussion and Conclusion}

In this paper, we have introduced \textit{ElegantRL-podracer}, a scalable and elastic library for cloud-native deep reinforcement learning. To efficiently utilize millions of GPU cores for DRL training, we first propose a tournament-based ensemble training scheme to orchestrate the training process on hundreds of GPUs, and then enable massively parallel simulation on thousands of GPU cores in a single GPU. Moreover, we follow the cloud-native paradigm to schedule the training of DRL agents by adhering to containerization, microservices, and MLOps. Thus, ElegantRL-podracer realizes the design principles in the respect of \textbf{scaling-out}, \textbf{efficiency}, and \textbf{accessibility}.

By presenting ElegantRL-podracer to the DRL community, we hope that ElegantRL-podracer can help address the data collection bottleneck using the manycore GPU architecture and apply DRL algorithms to complex real-world problems.

%% file: main.bbl
\begin{thebibliography}{31}
\providecommand{\natexlab}[1]{#1}
\providecommand{\url}[1]{\texttt{#1}}
\expandafter\ifx\csname urlstyle\endcsname\relax
  \providecommand{\doi}[1]{doi: #1}\else
  \providecommand{\doi}{doi: \begingroup \urlstyle{rm}\Url}\fi

\bibitem[Balalaie et~al.(2016)Balalaie, Heydarnoori, and
  Jamshidi]{Balalaie2016MicroservicesAE}
Armin Balalaie, A.~Heydarnoori, and Pooyan Jamshidi.
\newblock Microservices architecture enables devops: Migration to a
  cloud-native architecture.
\newblock \emph{IEEE Software}, 33:\penalty0 42--52, 2016.

\bibitem[Bernstein(2014)]{bernstein2014containers}
David Bernstein.
\newblock Containers and cloud: From lxc to docker to kubernetes.
\newblock \emph{IEEE Cloud Computing}, 1\penalty0 (3):\penalty0 81--84, 2014.

\bibitem[Brockman et~al.(2016)Brockman, Cheung, Pettersson, Schneider,
  Schulman, Tang, and Zaremba]{brockman2016openai}
Greg Brockman, Vicki Cheung, Ludwig Pettersson, Jonas Schneider, John Schulman,
  Jie Tang, and Wojciech Zaremba.
\newblock {OpenAI Gym}.
\newblock \emph{arXiv preprint arXiv:1606.01540}, 2016.

\bibitem[Castro et~al.(2018)Castro, Moitra, Gelada, Kumar, and
  Bellemare]{castro2018dopamine}
Pablo~Samuel Castro, Subhodeep Moitra, Carles Gelada, Saurabh Kumar, and Marc~G
  Bellemare.
\newblock Dopamine: A research framework for deep reinforcement learning.
\newblock \emph{arXiv preprint arXiv:1812.06110}, 2018.

\bibitem[Choquette et~al.(2021)Choquette, Gandhi, Giroux, Stam, and
  Krashinsky]{choquette2021nvidia}
Jack Choquette, Wishwesh Gandhi, Olivier Giroux, Nick Stam, and Ronny
  Krashinsky.
\newblock {NVIDIA A100 tensor core GPU}: Performance and innovation.
\newblock \emph{IEEE Micro}, 41\penalty0 (2):\penalty0 29--35, 2021.

\bibitem[Dhariwal et~al.(2017)Dhariwal, Hesse, Klimov, Nichol, Plappert,
  Radford, Schulman, Sidor, Wu, and Zhokhov]{baselines}
Prafulla Dhariwal, Christopher Hesse, Oleg Klimov, Alex Nichol, Matthias
  Plappert, Alec Radford, John Schulman, Szymon Sidor, Yuhuai Wu, and Peter
  Zhokhov.
\newblock Open{AI} baselines.
\newblock \url{https://github.com/openai/baselines}, 2017.

\bibitem[Fujimoto et~al.(2018)Fujimoto, Hoof, and
  Meger]{Fujimoto2018AddressingFA}
Scott Fujimoto, Herke Hoof, and David Meger.
\newblock Addressing function approximation error in actor-critic methods.
\newblock In \emph{International Conference on Machine Learning}, pages
  1587--1596. PMLR, 2018.

\bibitem[Gannon et~al.(2017)Gannon, Barga, and
  Sundaresan]{Gannon2017CloudNativeA}
Dennis Gannon, R.~Barga, and Neel Sundaresan.
\newblock Cloud-native applications.
\newblock \emph{IEEE Cloud Comput.}, 4:\penalty0 16--21, 2017.

\bibitem[Haarnoja et~al.(2018)Haarnoja, Zhou, Abbeel, and
  Levine]{Haarnoja2018SoftAO}
Tuomas Haarnoja, Aurick Zhou, P.~Abbeel, and Sergey Levine.
\newblock Soft actor-critic: Off-policy maximum entropy deep reinforcement
  learning with a stochastic actor.
\newblock In \emph{ICML}, 2018.

\bibitem[Herbst et~al.(2013)Herbst, Kounev, and
  Reussner]{Herbst2013ElasticityIC}
N.~Herbst, Samuel Kounev, and Ralf~H. Reussner.
\newblock Elasticity in cloud computing: What it is, and what it is not.
\newblock In \emph{ICAC}, 2013.

\bibitem[Hessel et~al.(2021)Hessel, Kroiss, Clark, Kemaev, Quan, Keck, Viola,
  and van Hasselt]{hessel2021podracer}
Matteo Hessel, Manuel Kroiss, Aidan Clark, Iurii Kemaev, John Quan, Thomas
  Keck, Fabio Viola, and Hado van Hasselt.
\newblock Podracer architectures for scalable reinforcement learning.
\newblock \emph{arXiv preprint arXiv:2104.06272}, 2021.

\bibitem[Li et~al.(2021)Li, Liu, Zheng, Wang, Walid, and
  Guo]{finrl_podracer_2021}
Zechu Li, Xiao-Yang Liu, Jiahao Zheng, Zhaoran Wang, Anwar Walid, and Jian Guo.
\newblock {FinRL-Podracer}: High performance and scalable deep reinforcement
  learning for quantitative finance.
\newblock \emph{ACM International Conference on AI in Finance (ICAIF)}, 2021.

\bibitem[Liang et~al.(2018)Liang, Liaw, Nishihara, Moritz, Fox, Goldberg,
  Gonzalez, Jordan, and Stoica]{liang2018rllib}
Eric Liang, Richard Liaw, Robert Nishihara, Philipp Moritz, Roy Fox, Ken
  Goldberg, Joseph Gonzalez, Michael Jordan, and Ion Stoica.
\newblock {RLlib}: Abstractions for distributed reinforcement learning.
\newblock In \emph{International Conference on Machine Learning}, pages
  3053--3062. PMLR, 2018.

\bibitem[Lillicrap et~al.(2016)Lillicrap, Hunt, Pritzel, Heess, Erez, Tassa,
  Silver, and Wierstra]{Lillicrap2016ContinuousCW}
Timothy~P Lillicrap, Jonathan~J Hunt, Alexander Pritzel, Nicolas Heess, Tom
  Erez, Yuval Tassa, David Silver, and Daan Wierstra.
\newblock Continuous control with deep reinforcement learning.
\newblock In \emph{ICLR}, 2016.

\bibitem[Liu et~al.(2020)Liu, Yang, Chen, Zhang, Yang, Xiao, and
  Wang]{liu2020finrl}
Xiao-Yang Liu, Hongyang Yang, Qian Chen, Runjia Zhang, Liuqing Yang, Bowen
  Xiao, and Christina~Dan Wang.
\newblock {FinRL}: A deep reinforcement learning library for automated stock
  trading in quantitative finance.
\newblock \emph{Deep Reinforcement Learning Workshop, NeurIPS}, 2020.

\bibitem[Liu et~al.(2021)Liu, Li, Wang, and Zheng]{elegantrl}
Xiao-Yang Liu, Zechu Li, Zhaoran Wang, and Jiahao Zheng.
\newblock {ElegantRL}: A lightweight and stable deep reinforcement learning
  library.
\newblock \url{https://github.com/AI4Finance-Foundation/ElegantRL}, 2021.

\bibitem[Makoviychuk et~al.(2021)Makoviychuk, Wawrzyniak, Guo, Lu, Storey,
  Macklin, Hoeller, Rudin, Allshire, Handa, et~al.]{makoviychuk2021isaac}
Viktor Makoviychuk, Lukasz Wawrzyniak, Yunrong Guo, Michelle Lu, Kier Storey,
  Miles Macklin, David Hoeller, Nikita Rudin, Arthur Allshire, Ankur Handa,
  et~al.
\newblock {Isaac Gym}: High performance {GPU}-based physics simulation for
  robot learning.
\newblock \emph{arXiv preprint arXiv:2108.10470}, 2021.

\bibitem[Mell and Grance(2011)]{Mell2011TheND}
P.~Mell and T.~Grance.
\newblock The nist definition of cloud computing.
\newblock In \emph{CSRC}, 2011.

\bibitem[Mnih et~al.(2013)Mnih, Kavukcuoglu, Silver, Graves, Antonoglou,
  Wierstra, and Riedmiller]{Mnih2013PlayingAW}
V.~Mnih, K.~Kavukcuoglu, D.~Silver, A.~Graves, Ioannis Antonoglou, Daan
  Wierstra, and Martin~A. Riedmiller.
\newblock Playing atari with deep reinforcement learning.
\newblock \emph{ArXiv}, abs/1312.5602, 2013.

\bibitem[{OpenAI}(2018)]{spinningup}
{OpenAI}.
\newblock {OpenAI} spinning up.
\newblock \url{https://spinningup.openai.com}, 2018.

\bibitem[Raffin et~al.(2019)Raffin, Hill, Ernestus, Gleave, Kanervisto, and
  Dormann]{raffin2019stable}
Antonin Raffin, Ashley Hill, Maximilian Ernestus, Adam Gleave, Anssi
  Kanervisto, and Noah Dormann.
\newblock {Stable baselines3}.
\newblock \emph{GitHub repository}, 2019.

\bibitem[Salimans et~al.(2017)Salimans, Ho, Chen, Sidor, and
  Sutskever]{salimans2017evolution}
Tim Salimans, Jonathan Ho, Xi~Chen, Szymon Sidor, and Ilya Sutskever.
\newblock Evolution strategies as a scalable alternative to reinforcement
  learning.
\newblock \emph{arXiv preprint arXiv:1703.03864}, 2017.

\bibitem[Schulman et~al.(2017)Schulman, Wolski, Dhariwal, Radford, and
  Klimov]{Schulman2017ProximalPO}
J.~Schulman, F.~Wolski, Prafulla Dhariwal, Alec Radford, and Oleg Klimov.
\newblock Proximal policy optimization algorithms.
\newblock \emph{ArXiv}, abs/1707.06347, 2017.

\bibitem[Service(2015)]{wrds}
Wharton Research~Data Service.
\newblock Standard \& {Poor’s} compustat, 2015.
\newblock Data retrieved from Wharton Research Data Service.

\bibitem[Silver et~al.(2017)Silver, Schrittwieser, Simonyan, Antonoglou, Huang,
  Guez, Hubert, Baker, Lai, Bolton, et~al.]{silver2017mastering}
David Silver, Julian Schrittwieser, Karen Simonyan, Ioannis Antonoglou, Aja
  Huang, Arthur Guez, Thomas Hubert, Lucas Baker, Matthew Lai, Adrian Bolton,
  et~al.
\newblock Mastering the game of {Go} without human knowledge.
\newblock \emph{Nature}, 550\penalty0 (7676):\penalty0 354--359, 2017.

\bibitem[Stooke and Abbeel(2019)]{stooke2019rlpyt}
Adam Stooke and Pieter Abbeel.
\newblock rlpyt: A research code base for deep reinforcement learning in
  pytorch.
\newblock \emph{arXiv preprint arXiv:1909.01500}, 2019.

\bibitem[system~reference architecture(2020)]{NVIDIA_SupPod2020}
NVIDIA DGX~A100 system~reference architecture.
\newblock \emph{{NVIDIA DGX SuperPOD}: Scalable infrastructure for AI
  leadership}.
\newblock NVIDIA Corporation, 2020.

\bibitem[Todorov et~al.(2012)Todorov, Erez, and Tassa]{todorov2012mujoco}
Emanuel Todorov, Tom Erez, and Yuval Tassa.
\newblock {MuJoCo}: A physics engine for model-based control.
\newblock In \emph{IEEE/RSJ International Conference on Intelligent Robots and
  Systems}, pages 5026--5033. IEEE, 2012.

\bibitem[Van~Hasselt et~al.(2016)Van~Hasselt, Guez, and
  Silver]{Hasselt2016DeepRL}
Hado Van~Hasselt, Arthur Guez, and David Silver.
\newblock Deep reinforcement learning with double {Q}-learning.
\newblock In \emph{Proceedings of the AAAI Conference on Artificial
  Intelligence}, volume~30, 2016.

\bibitem[Wijmans et~al.(2020)Wijmans, Kadian, Morcos, Lee, Essa, Parikh, Savva,
  and Batra]{Wijmans2020DDPPOLN}
Erik Wijmans, Abhishek Kadian, Ari~S. Morcos, Stefan Lee, Irfan Essa, Devi
  Parikh, M.~Savva, and Dhruv Batra.
\newblock {DD-PPO}: Learning near-perfect pointgoal navigators from 2.5 billion
  frames.
\newblock In \emph{ICLR}, 2020.

\bibitem[Zhang and Mo(2021)]{zhang2021reinforcement}
Tengteng Zhang and Hongwei Mo.
\newblock Reinforcement learning for robot research: A comprehensive review and
  open issues.
\newblock \emph{International Journal of Advanced Robotic Systems}, 18\penalty0
  (3):\penalty0 17298814211007305, 2021.

\end{thebibliography}
